\documentclass[10pt,letterpaper]{article}
\usepackage[top=0.85in,left=2.75in,footskip=0.75in]{geometry}
\usepackage{booktabs}

\usepackage{amsmath,amssymb}
\usepackage[inkscapelatex=false]{svg}

\usepackage{changepage}

\usepackage{textcomp,marvosym}

\usepackage{cite}

\usepackage{nameref,hyperref}

\usepackage[right]{lineno}

\usepackage[nopatch=eqnum]{microtype}
\DisableLigatures[f]{encoding = *, family = * }

\usepackage{array}

\newcolumntype{+}{!{\vrule width 2pt}}

\newlength\savedwidth

\raggedright
\usepackage[aboveskip=1pt,labelfont=bf,labelsep=period,justification=raggedright,singlelinecheck=off]{caption}

\makeatletter
\renewcommand{\@biblabel}[1]{\quad#1.}
\makeatother

\usepackage{lastpage,fancyhdr,graphicx}
\usepackage{epstopdf}
\fancyheadoffset[L]{2.25in}
\fancyfootoffset[L]{2.25in}
\begin{document}
\vspace*{0.2in}

\begin{flushleft}
{\Large
\textbf\newline{Open-source data pipeline for street-view images: a case study on community mobility during COVID-19 pandemic} 
}
\newline
\\
Matthew Martell\textsuperscript{1*\Yinyang},
Nick Terry\textsuperscript{1\Yinyang},
Ribhu Sengupta\textsuperscript{1},
Chris Salazar\textsuperscript{1},
Nicole A. Errett\textsuperscript{2},
Scott B. Miles\textsuperscript{3},
Joseph Wartman\textsuperscript{4},
Youngjun Choe\textsuperscript{1},
\\
\bigskip
\textbf{1} Industrial \& Systems Engineering, University of Washington, Seattle, WA, United States
\\
\textbf{2} Environmental \& Occupational Health Sciences, University of Washington, Seattle, WA, United States
\\
\textbf{3} Human Centered Design \& Engineering, University of Washington, Seattle, WA, United States
\\
\textbf{4} Civil \& Environmental Engineering, University of Washington, Seattle, WA, United States
\\
\bigskip

%
%
\Yinyang These authors contributed equally to this work.





* corresponding author marte292@uw.edu

\end{flushleft}
\section*{Abstract}
Street View Images (SVI) are a common source of valuable data for researchers. Researchers have used SVI data for estimating pedestrian volumes, demographic surveillance, and to better understand built and natural environments in cityscapes. However, the most common source of publicly available SVI data is Google Street View. Google Street View images are collected infrequently, making temporal analysis challenging, especially in low population density areas. Our main contribution is the development of an open-source data pipeline for processing 360-degree video recorded from a car-mounted camera. The video data is used to generate SVIs, which then can be used as an input for temporal analysis. We demonstrate the use of the pipeline by collecting a SVI dataset over a 38-month longitudinal survey of Seattle, WA, USA during the COVID-19 pandemic. The output of our pipeline is validated through statistical analyses of pedestrian traffic in the images. We confirm known results in the literature and provide new insights into outdoor pedestrian traffic patterns. This study demonstrates the feasibility and value of collecting and using SVI for research purposes beyond what is possible with currently available SVI data. Limitations and future improvements on the data pipeline and case study are also discussed.

\section*{Introduction}

Street-level imagery is becoming an increasingly popular form of data for research \cite{cinnamon_panoramic_2021}. In particular, Street View Images (SVI) as popularized by Google Street View are used in many studies \cite{gilge_google_2016,li_street_2022}. Uses for SVI data include estimating demographics \cite{GebruTimnit2017Udla}, evaluating the built environment \cite{badland_can_2010}, surveying plant species \cite{ringland_automated_2021}, measuring pedestrian volume \cite{yin_big_2015} and many other applications \cite{rzotkiewicz_systematic_2018,wang_using_2019, novack_towards_2020}. 

While SVI data can provide many useful insights for researchers, it is not without its flaws. For corporate-collected images such as Google Street View, or Tencent Street View the availability of images depends on where the companies decide to collect data, while the accessibility of these images hinges on the companies' data provision policies. For example, there is no Google Street View service in most parts of Africa. An alternative to corporate-collected images are crowdsourced SVI databases such as Mapillary \cite{mapillary}. These crowdsourced images sometimes will have better coverage or temporal resolution than Google Street View, at the cost of varying image quality, field of view, and positional accuracy \cite{li_street_2022,mahabir_crowdsourcing_2020}. Perhaps the largest challenge with SVI data is its temporal instability. Updates to these image datasets at specific locations are infrequent, especially in rural areas \cite{cinnamon_panoramic_2021,suel_multimodal_2021, smith_google_2021}. Additionally, images frequently are not collected at a consistent time of day, or season, even within the same city. These issues make existing SVI data unreliable for temporal studies. 

Typically, temporal studies involving image data use images (or video) from fixed locations. This data is used to do things such as evaluate disaster recovery \cite{burton_evaluating_2011}, monitor ecological change \cite{depauw_use_2022}, or measure urban flooding \cite{hao_estimating_2022}. Data from fixed cameras is also used to count people \cite{velastin_detecting_2020}. The challenge with these methods is that they are fixed-location. In order to collect spatial image data for these methods, frequently a large team is required to traverse areas on foot. This challenge, along with existing SVI data's temporal issues demonstrate the potential value of collecting longitudinal SVI data.   

Our main contribution is demonstrating the feasibility of collecting longitudinal SVI data. We demonstrate this through the creation of a complete data pipeline for conducting pedestrian counts using car-based street-level imagery. The pipeline accepts raw video collected by the camera as an input and outputs a record of each pedestrian detection and their locations (latitude and longitude). This approach allows for analysis of mobility patterns with high spatial resolution and a short lag time. 

Specifically, we use this pipeline to generate and analyze video from 37 video-collection runs in the city of Seattle, Washington, USA from May 2020 through July 2023. The video data was converted into over 4 million high-resolution images, with each data-collection run representing about 1.5 TB of image data. We used the images to create a record containing the location of each detected pedestrian, cross-referenced to the relevant GEOID \cite{census_geoid}. To detect pedestrians in the still images, our pipeline leverages the state-of-the-art convolutional neural network, Pedestron \cite{hasan2022pedestrian}. We used the cascade{\_}hrnet architecture benchmarked on on the CrowdHuman data set \cite{shao2018crowdhuman}.

As a secondary contribution, we provide a case study based on the video data collected throughout the COVID-19 pandemic. We examine the effect of vaccine availability and local demographics on pedestrian detections, while accounting for weekly and yearly seasonality. Our findings demonstrate the utility of our data processing pipeline in tracking community mobility over time and show the potential for its use in a variety of research domains.

\section*{Methods}
\subsection*{Data Collection}

We collected our data as a part of the the Seattle street-level imagery campaign, an ongoing series of video surveys for the purposes of documenting mobility throughout the COVID-19 pandemic. During each survey, a vehicle equipped with a 360$^\circ$ video camera is driven along pre-defined route through Seattle while collecting video data and GPS metadata. The route incorporates broad neighborhood/area canvassing designed to collect data useful to multidisciplinary researchers as well as capital transects. Full details on the route design are available in Errett et al. \cite{errett2023street}. The capital transects specifically target capitals (social, cultural, built, economic, and public health) which are theorized to be closely tied to community resilience \cite{miles2015foundations}. Specific canvassing areas and capitals within Seattle were chosen to ensure a representative sample of the overall population of Seattle \cite{errett2023street}. While the drivers try to make the surveys as consistent as possible, occasionally exogenous factors caused deviations from standard protocols. For example, during three of the surveys (05-29-2020, 06-18-2020, and 06-26-2020), protests over the murder of George Floyd caused parts of the survey route to be unnavigable. 

\subsection*{Data Processing Pipeline}

After video collection, the raw data is segmented into image data. The images are sub-sampled from video frames so that they are collected about every 4 meters. The images are then uploaded into the DesignSafe-CI Data Depot\cite{RathjeEllenM2017DNCf}. From DesignSafe, the images are transferred to the TACC Frontera high-performance computing cluster \cite{frontera}. We completed all file transfers between the two services using Globus \cite{Globus}. Without access to these services, or similar ones, the storage and computing requirements for this project would be intractable. 

\begin{figure}[!h]
\includegraphics[width=\textwidth]{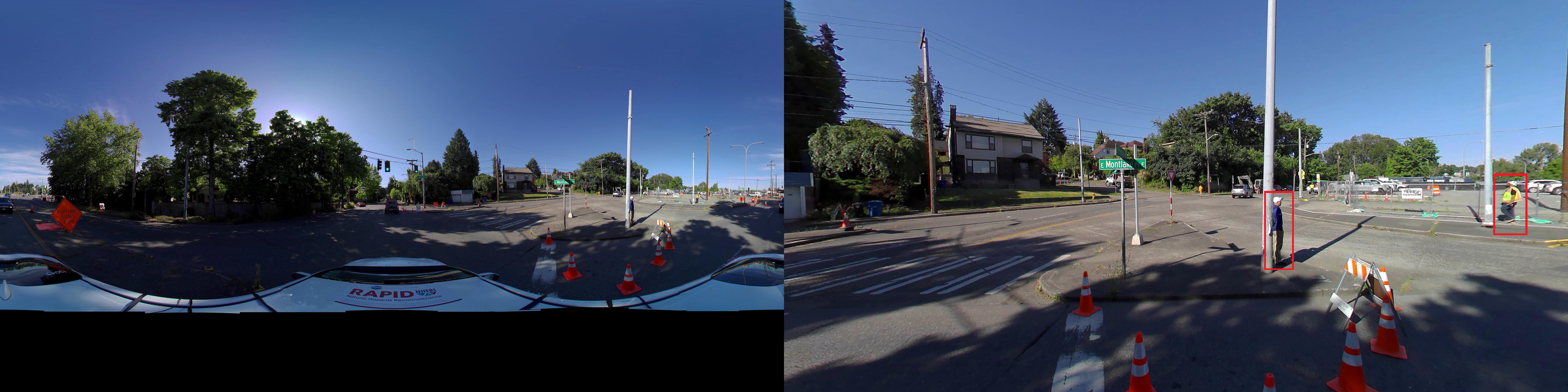}
\caption{Sample images from the pedestrian detection data pipeline. The left image is an original 360$^\circ$ image from a data collection run. The image on the right is the right-hand side of the original image  after orthorectification and pedestrian detection (both sides of the image are processed separately). There are two pedestrians that were detected by the algorithm (in red bounding boxes).}

\label{fig:example}
\end{figure}

On Frontera, orthorectification is performed to the images, then pedestrian detection is performed on the orthorectified images. The orthorectification transforms the images from a single image in the equirectangular projection to two images in the rectilinear (gnomonic) projection \cite{Orthorecti}. Pedestrians are detected on each of the new images using a convolutional neural network (CNN) based on a pre-trained model from the Pedestron repository \cite{hasan2022pedestrian}. Our data represents a highly challenging detection task, as there is great variation in lighting, backgrounds, human poses, levels of occlusion and crowd density from image to image and run to run. The Cascade Mask R-CNN architecture in the Pedestron repository performed well on the CrowdHuman data set, representing a similar challenge to our data \cite{shao2018crowdhuman}. All testing and use of the CNN was performed using GPUs on the Frontera cluster. An example image after undergoing orthorectification and pedestrian detection is shown in Fig \ref{fig:example}.  

Using one GPU node on Frontera, with four NVIDIA Quadro RTX 5000 GPUs, the entire process takes about 3 seconds per original 360$^\circ$ image. Given the 4 million images we collected, this takes about 3,300 hours of computing time. While this is not a small number, when running in parallel, the whole process can be completed in a manner of days. In comparison, a human taking 10s per orthorectified image to count all the pedestrians would take over 22,000 hours to complete the same task. File compression/decompression for file transfer also takes a substantial amount of time. Since we used DesignSafe as our main data storage platform, we had to transfer files to/from the Frontera supercomputer to perform our pedestrian detection. To avoid overloading the file transfer system, we compressed the images from each run into a tar file prior to transferring the files to Frontera. This file compression/decompression can take several hours per run, but can be performed in parallel with the detection algorithm since they are on different systems. After compression, file transfer using Globus \cite{Globus} takes minutes. 

In post-processing, the pipeline filters out low-confidence detections (defined as any detection with less than $80\%$ confidence) and associates the remaining high-confidence detections to U.S. Cenus Bureau GEOIDs \cite{census_geoid}. We arrived at this confidence level after tuning for the precision and recall of the CNN classifier. Specifically, the pipeline filters based on the output of the second to last layer of the CNN, known as a \textit{softmax layer}. For a $k-$class classification problem, the softmax layer will output a $k-$dimensional probability vector, where each $i^{th}$ entry of the vector gives the probability that the original input to the CNN belongs to class $i.$

The final stage of post-processing is GEOID matching, where latitude and longitude metadata are cross-referenced to disjoint geographic regions (e.g. U.S. census tracts or block groups) and their respective GEOID codes. The cross-referencing code assumes the availability of shapefiles describing the geometry of the geographic regions. Aggregating the pedestrian detections according to U.S. Census Bureau GEOIDs \cite{census_geoid} is necessary for analyses using sociodemographic data collected by the census. Additionally, the pedestrian detections can easily be cross-referenced with custom geometry defined using popular geographic information system software, such as the capitals data used in route construction and our analysis.

\begin{figure}[!h]
\includegraphics[width=\textwidth]{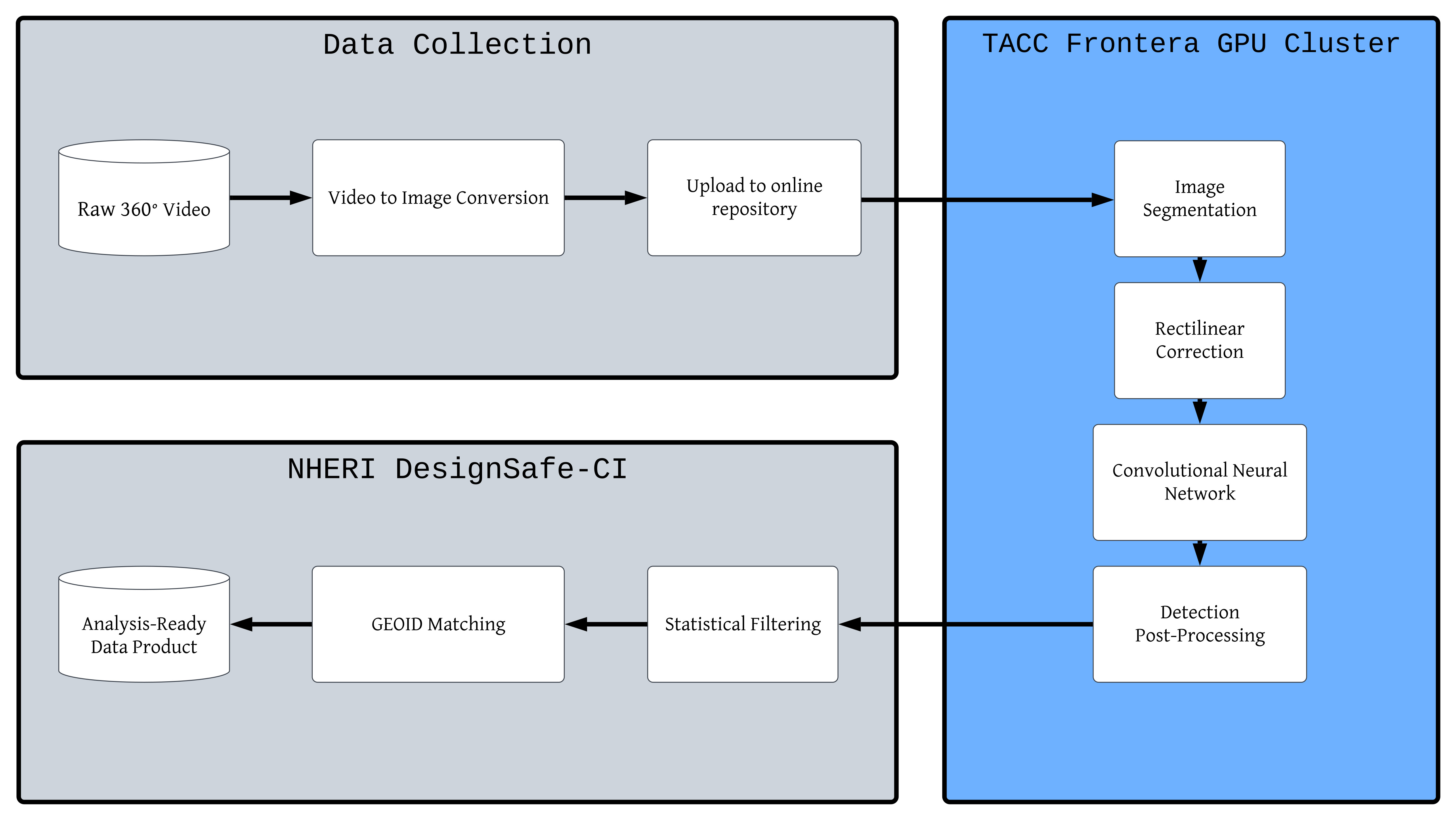}
\caption{Flowchart of the data processing pipeline. The parts of the flowchart in gray occur on NHERI DesignSafe-CI, while the right hand part in blue is done on the Frontera cluster.}
\label{fig:pipeline}
\end{figure}

Following the GEOID matching step, the pedestrian detections data is written to a tabular format file (e.g. comma separated values). This file is an ``analysis-ready'' data product, in the sense that it is readable by most popular statistical analysis software (R, SPSS, Stata, etc.) and can be easily merged with other datasets using the GEOID column(s). A visual depiction of the entire pipeline is seen in Fig \ref{fig:pipeline}. Full code and a manual for following our process is available at \url{https://github.com/marte292/rapid-data-pipeline}.

\subsection*{Case Study: Community Mobility in Seattle during the COVID-19 Pandemic}

\subsubsection*{Data Processing}
All analysis is performed using the Python programming language version 3.11 \cite{Python}. The initial data product as outlined in the previous section is a list of detections, alongside the date of collection, geolocation, and GEOID. We also utilized a similar list of the images themselves with the same features. The last dataset we utilized is the median household income data and racial demographic data from the 2019 American Community Survey (ACS) 5-year estimates. We aggregated the detections and image data for each data collection survey at the census tract level, then matched each census tract's total number of detections and images to its respective demographic and income data. 

We utilized the data from 36 of the 37 surveys, omitting data from 10-29-2020. A heavy rain event caused the survey to be stopped early due to poor video quality. For each survey, we divided the number of detections in each census tract by the number of images collected in the tract to create a normalized `detections per image' metric. This is a necessary step as the number of images in each tract may change survey to survey due to circumstances outside our control, such as construction or community events altering the route. 

The last step in data processing was to transform some of our data to be represented by categorical variables. The date of each survey was coded both as either a weekend or weekday, and by the season. The date was also coded as either being before, or after the date that vaccines became publicly available. Income data was coded to be one of 5 levels that were used during route design. Lastly, the proportion of the census tract's population that identifies as non-white was coded as an indicator variable with '1' corresponding to areas that are $55.5\%$ white or more. We determined this threshold using Jenk's natural breaks optimization. This left us with a dataset of 3171 observations to be used for analysis. Each observation represented a census tract with a detections per image value, as well as values for each of the categorical variables defined above.

\subsubsection*{Initial Regression Analysis}

Based on the known literature, we hypothesized that season, day of the week, COVID-19 vaccine availability, income level, and demographics all would have an impact on pedestrian traffic. We implemented a regression model to identify which of these factors are identified as statistically significant ($\alpha=.05$). The regression model is detailed below:

\begin{equation} \label{eq:1}
\begin{aligned}
Y = \beta_{0}+\beta_{1} \times I_{vaccine} +\beta_{2..4} \times C_{season} +\beta_{5} \times I_{weekend} \\
+\beta_{6..9} \times C_{income level} +\beta_{10} \times I_{demographic indicator} +\epsilon,
\end{aligned}
\end{equation}
where $Y$ is the detections per image for a date/census tract combination; $I_{vaccine}$ is an indicator for if the vaccine was available on that date; $C_{season}$ is a categorical variable with 3 levels for summer, winter, and spring; $I_{weekend}$ is an indicator for if it is the weekend or not; $C_{income level}$ is a categorical variable with 4 levels for the 4 income brackets above the lowest bracket; $I_{demographic indicator}$ is an indicator variable for if the population is $55.5\%$ white or more. $\beta_{0}$ is the baseline detections per image on a weekday, not in the summer, with the vaccine unavailable, in a census tract at the lowest income level and a population less than $55.5\%$ white. $\beta_{1}$ represents the change in detections per image from the vaccine becoming available, and $\beta_{2..4}$ represent the change  for different seasons. $\beta_{5}$ represents the change from a weekday to the weekend, and $\beta_{6..9}$ represent the change to other income brackets. Lastly $\beta_{10}$ represents the change in detections per image to from an area that is less than $55.5\%$ white an area that is more.

In addition to the above analysis, we subset the data by only looking at detections that occurred in an image with at least one other detection. Then we calculated detections per image again, and fit the above model again with the new response variable. This same process was followed for detections with at least two, three, and four other detections in the same image. The goal of these analysis was to see if there were different trends for larger groups of people when compared with the entire data set.

\section*{Results}

\subsection*{Data pipeline} Our main contribution, the open-source data pipeline, is publicly available on \url{https://github.com/marte292/rapid-data-pipeline}. The repository contains a process manual with step-by-step instructions on how to implement the data pipeline in Python \cite{Python}. The required Python libraries and system requirements are provided. Additionally, we provide enough code for future researchers to implement the pipeline on their own systems, with their own file structure. The pipeline is capable of processing terabytes of image data and outputting an analysis-ready data product in a matter of days (using high-performance computing, such as a single GPU node on Frontera, an academic supercomputer) with minimal human input. 

\subsection*{Case study}

\begin{figure}[!h]
\includegraphics[width=\textwidth]{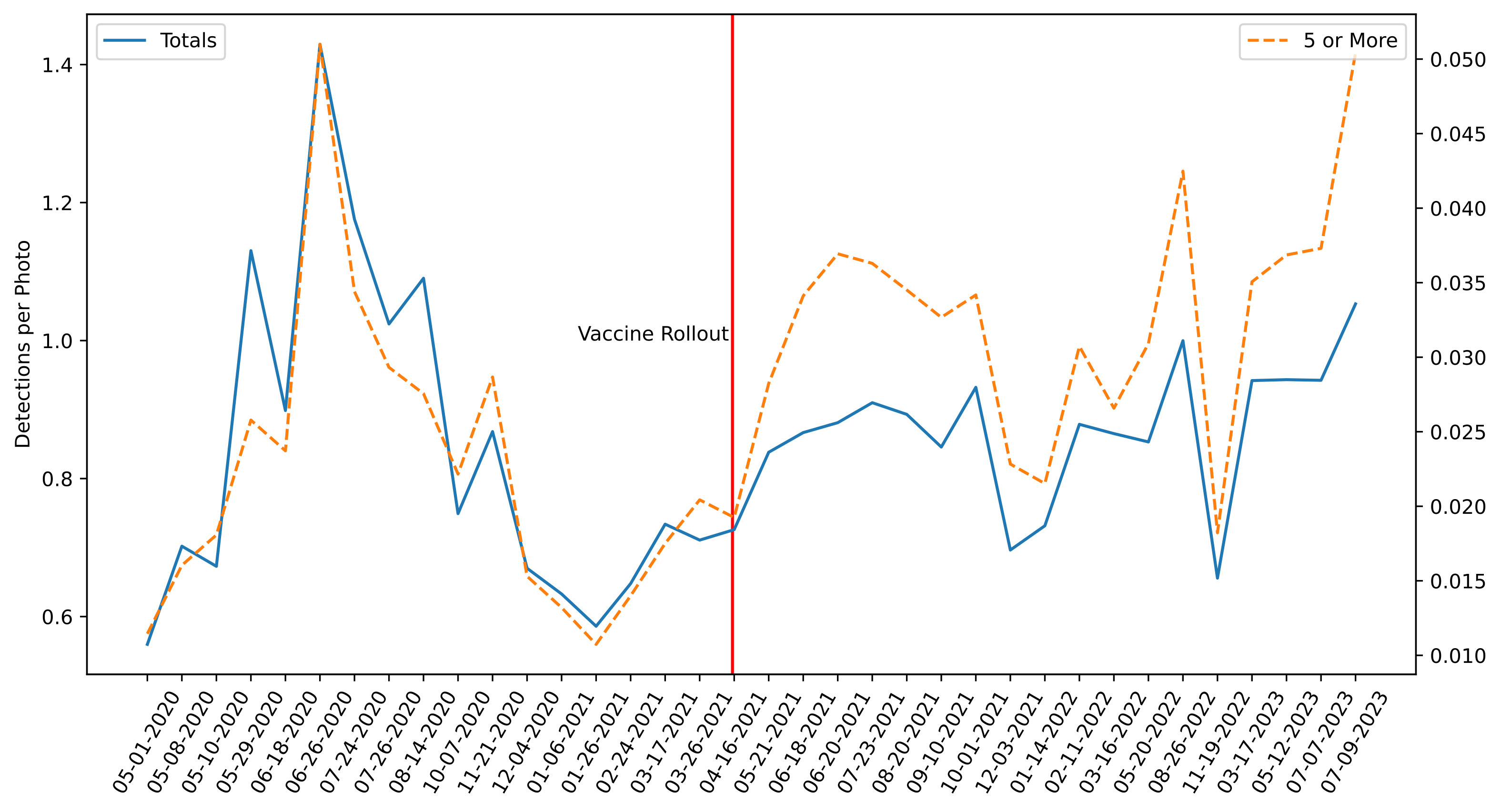}
\caption{Time series data of the total detections per image (solid blue line, left axis), and detections per image for the subset of detections sharing an image with at least 4 others (orange dashed line, right axis). As the survey dates are irregular, all dates are included in the figure. Please note that the axis for total detections per image does not start at 0. This was done purposefully to facilitate comparison between the trends of the two graphs.}
\label{fig:timeseries}
\end{figure}

Using data from the Seattle street-level imagery campaign, we calculated the number of detections per image across all data collection surveys. Fig \ref{fig:timeseries} shows the detections per image for each survey, as well as the detections per image for the subset of detections sharing an image with at least 4 others. Fig \ref{fig:timeseries} also displays the timestamp of COVID-19 vaccines becoming publicly available in Washington state. 

Fig \ref{fig:timeseries} depicts the trends over time for detections per image and detections sharing an image with at least 4 others. While both graphs exhibit similar trends overall, notably after vaccine rollout the graph of detections sharing an image with at least 4 others exceeds the graph of detections per image in all cases. The spike in detections seen in June 2020 is due to the large scale protests of police brutality that took place in Seattle in the aftermath of George Floyd's murder.

The full results of the linear regression model for total detections per image are displayed in Table \ref{tab:regression1}. They show that the season being summer is the only significant seasonal effect. Additionally, the income bracket is a significant predictor, with wealthier areas seeing less pedestrian traffic. Finally, a census tract having a population greater than $55.5\%$ white is a significant positive predictor. All other variables are not significant, including vaccine availability. 

\begin{table} 
\begin{tabular}{lclc} 
\toprule
\textbf{Dep. Variable:}            & Detections\_per\_Image & \textbf{  R-squared:         } &     0.086   \\
\textbf{Model:}                    &          OLS           & \textbf{  Adj. R-squared:    } &     0.083   \\
\textbf{Method:}                   &     Least Squares      & \textbf{  F-statistic:       } &     29.71   \\
\textbf{No. Observations:}         &           3171    & \textbf{  Prob (F-statistic):} &  3.20e-55   \\
\textbf{Df Residuals:}             &           3160        & \textbf{  Log-Likelihood:    } &   -4456.8   \\
\textbf{Df Model:}                 &             10         & \textbf{                     } &             \\
\textbf{Covariance Type:}          &       nonrobust        & \textbf{                     } &             \\
\bottomrule
\end{tabular}
\begin{tabular}{lcccccc}
                                   & \textbf{coef} & \textbf{std err} & \textbf{t} & \textbf{P$> |$t$|$} & \textbf{[0.025} & \textbf{0.975]}  \\
\midrule
\textbf{Intercept}                 &       1.1242  &        0.075     &    14.914  &         0.000        &        0.976    &        1.272     \\
\textbf{Spring}          &       0.0783  &        0.053     &     1.471  &         0.141        &       -0.026    &        0.183     \\
\textbf{Summer}          &       0.2527  &        0.055     &     4.599  &         0.000        &        0.145    &        0.360     \\
\textbf{Winter}          &      -0.0046  &        0.058     &    -0.079  &         0.937        &       -0.118    &        0.109     \\
\textbf{Vaccine Available}            &       0.0061  &        0.036     &     0.172  &         0.863        &       -0.064    &        0.076     \\
\textbf{Weekend}            &      -0.0775  &        0.052     &    -1.483  &         0.138        &       -0.180    &        0.025     \\
\textbf{Income Bracket 2} &      -0.4689  &        0.081     &    -5.795  &         0.000        &       -0.628    &       -0.310     \\
\textbf{Income Bracket 3} &      -0.8688  &        0.079     &   -11.041  &         0.000        &       -1.023    &       -0.714     \\
\textbf{Income Bracket 4} &      -0.9938  &        0.086     &   -11.540  &         0.000        &       -1.163    &       -0.825     \\
\textbf{Income Bracket 5} &      -1.3752  &        0.116     &   -11.893  &         0.000        &       -1.602    &       -1.148     \\
\textbf{More than 55.5\% White}         &       0.6416  &        0.054     &    11.820  &         0.000        &        0.535    &        0.748     \\
\bottomrule

\end{tabular}
\caption{OLS Regression Results for Detections per Image.}
\label{tab:regression1}
\end{table}

 For the regression models using a subset of data, the results are similiar to the initial model. All models have the same significant predictors as the initial model. The model using the detections sharing an image with at least one other also had the weekend as a borderline significant, negative predictor. The models using detections sharing an image with at least 3 and 4 others had vaccine availability as a significant, positive predictor. The full results of the linear regression model for detections per image with at least 4 others are displayed in Table \ref{tab:regression2}, with all other regression models available in the supporting information.

\begin{table} 
\begin{tabular}{lclc}
\toprule
\textbf{Dep. Variable:}            & Five\_Or\_More\_Peds\_per\_Image & \textbf{  R-squared:         } &     0.059   \\
\textbf{Model:}                    &               OLS                & \textbf{  Adj. R-squared:    } &     0.056   \\
\textbf{Method:}                   &          Least Squares           & \textbf{  F-statistic:       } &     19.78   \\
\textbf{No. Observations:}         &                3171              & \textbf{  Prob (F-statistic):} &  7.09e-36   \\
\textbf{Df Residuals:}             &                3160              & \textbf{  Log-Likelihood:    } &    4286.3   \\
\textbf{Df Model:}                 &                  10              & \textbf{                     } &             \\
\textbf{Covariance Type:}          &            nonrobust             & \textbf{                     } &             \\
\bottomrule
\end{tabular}
\begin{tabular}{lcccccc}
                                   & \textbf{coef} & \textbf{std err} & \textbf{t} & \textbf{P$> |$t$|$} & \textbf{[0.025} & \textbf{0.975]}  \\
\midrule
\textbf{Intercept}                 &       0.0393  &        0.005     &     8.213  &         0.000        &        0.030    &        0.049     \\
\textbf{Spring}          &       0.0006  &        0.003     &     0.180  &         0.857        &       -0.006    &        0.007     \\
\textbf{Summer}          &       0.0100  &        0.003     &     2.870  &         0.004        &        0.003    &        0.017     \\
\textbf{Winter}          &      -0.0039  &        0.004     &    -1.057  &         0.291        &       -0.011    &        0.003     \\
\textbf{Vaccine Available}            &       0.0093  &        0.002     &     4.109  &         0.000        &        0.005    &        0.014     \\
\textbf{Weekend}            &       0.0009  &        0.003     &     0.270  &         0.787        &       -0.006    &        0.007     \\
\textbf{Income Bracket 2} &      -0.0289  &        0.005     &    -5.626  &         0.000        &       -0.039    &       -0.019     \\
\textbf{Income Bracket 3} &      -0.0468  &        0.005     &    -9.373  &         0.000        &       -0.057    &       -0.037     \\
\textbf{Income Bracket 4} &      -0.0541  &        0.005     &    -9.889  &         0.000        &       -0.065    &       -0.043     \\
\textbf{Income Bracket 5} &      -0.0688  &        0.007     &    -9.376  &         0.000        &       -0.083    &       -0.054     \\
\textbf{More than 55.5\% White}         &       0.0309  &        0.003     &     8.960  &         0.000        &        0.024    &        0.038     \\
\bottomrule
\end{tabular}
\caption{OLS Regression Results for Detections per Image for the detections subset sharing an image with at least 4 others.}
\label{tab:regression2}
\end{table}

\section*{Discussion}
\subsection*{Comparison to Google Community Mobility Data}

Given the ability to measure community mobility through pedestrian counts, there is potential value of our pipeline for social sciences and public health research \cite{Hou2021, schlosser2020}. At an individual level, higher physical activity is known to predict better physical \cite{baker2003measuring,petersen2015time} and mental health \cite{zhu2018daily,polku2015life,vallee2011role}, and is associated with higher self-reported satisfaction and quality of life \cite{mulry2019relationship,bergstad2011subjective}. In an aggregate sense, mobility is theorized to be an intermediate variable through which socioeconomic deprivation affects vulnerability to infectious disease \cite{ossimetha2021socioeconomic,zhai2021social}, resilience to disasters \cite{hong2021measuring}, and exposure to environmental hazards \cite{lewis2019exposures}. In light of this body of literature, we argue that the use of pedestrian counts to assess mobility could be a differentiating factor in researching social and health inequity. One extremely common source of mobility data during the COVID-19 Pandemic has been Google Community Mobility Reports \cite{Google} and Apple Mobility Trends Reports \cite{Apple}. While there have been improvements in recent years \cite{stockham_causal_2022}, there are known representation and self-selection biases with existing mobility data captured by smartphones and other internet-based data collection methods \cite{jonas_klingwort_critical_2020,liu_physical_2022, roy_2019, milusheva2021assessing, aleta}. 

Given the large number of publications using smartphone data as the foundation for their work, a natural question is how our data compares to smartphone mobility data. Comparison between our data set and the still publicly available Google Community Mobility Reports data  can reveal some of the similarities and differences between the two data sets \cite{Google}. Google Community Mobility data is reported at the county level in the United States. Since Seattle is in King County, Washington, the King County data is what we use to draw the comparison. 

Google Community Mobility data does not provide raw mobility numbers, but rather is reported as a percentage change from the five week period of Jan 5 - Feb 6, 2020. This data is collected from smartphones running the Android operating system with location history turned on, which is off by default. The data is baselined by day of week, so data from a given Monday is compared to the median of the five Mondays in the baseline window to calculate a percent change. Additionally, it is unclear how exactly Google quantifies mobility. It is mentioned that it combines number of visitors to a location with amount of time spent in that location, but no specifics beyond that are provided. 

Google mobility data is broken down into different categories. The category that most closely aligns with one of the categories used in our analysis is parks. Although Google's data classifies parks as official national parks and not the general outdoors, it does not indicate how it accounts for city or state parks. Our own data for park locations is based on the City of Seattle's official classifications. 

\begin{figure*}[h!]
    \centering
    \includegraphics[width=\textwidth]{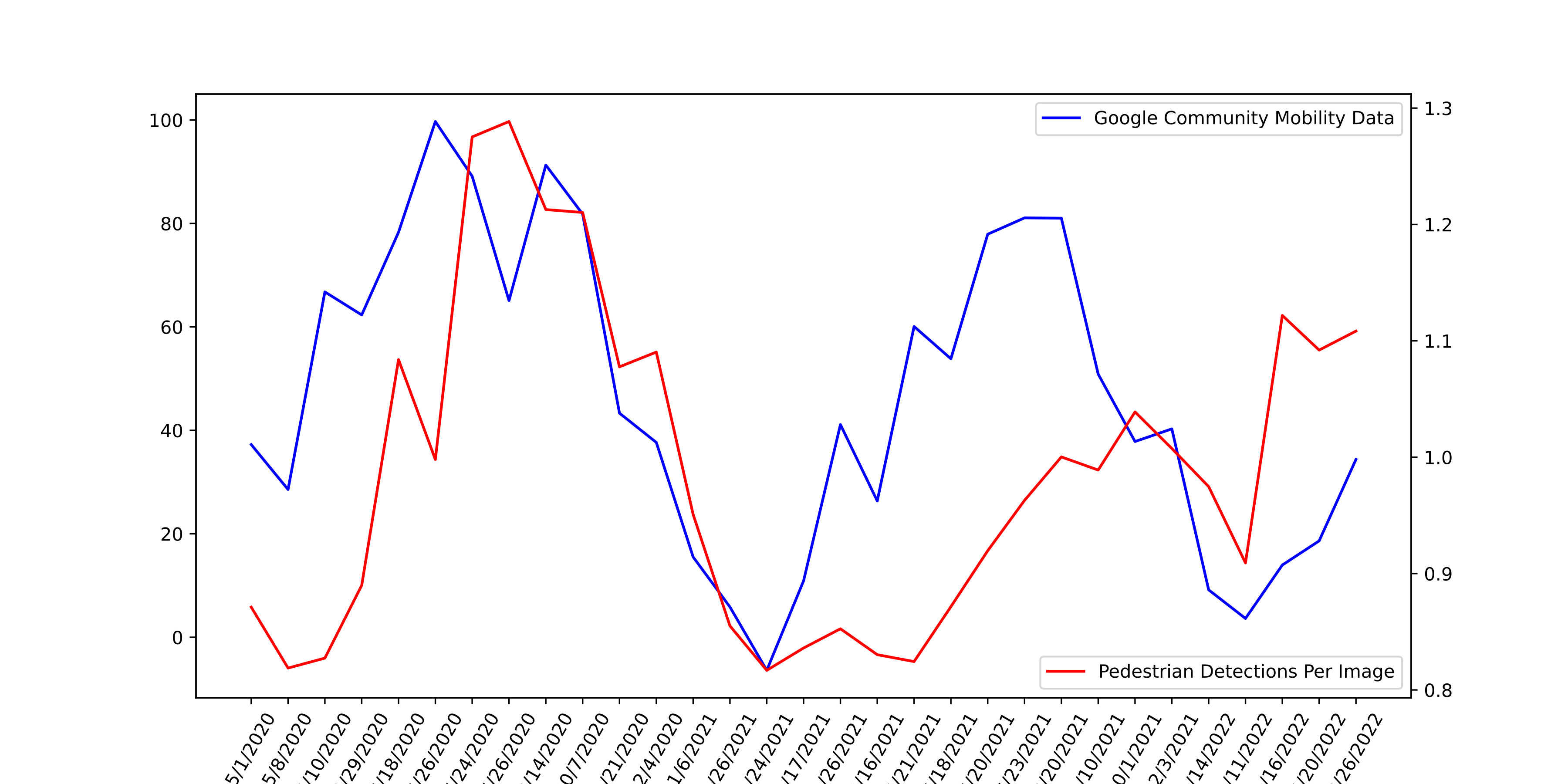}{}
    \caption{A depiction of our own detections per image data (blue, dashed; right axis) against Google Community Mobility data (orange, solid; left axis). The pearson correlation between the two data sets is 0.387. The Google Community mobility data is aggregated at King County, WA, while our data covers a survey route within Seattle, which belongs to King County. As the dates of surveys were irregular (e.g., due to weather conditions), all dates are included in the figure.}
    \label{fig:Google}
\end{figure*}

Fig \ref{fig:Google} shows a comparison of our detections per image data against Google Community Mobility data. Note that not all surveys are included because Google Community Mobility data stopped being provided on October 15, 2022. Overall, the trends between the two data sets are remarkably similar, lending further credibility to our data collection procedure. The more notable differences in the graph are from the months of November 2020 through August 2021, where the Google mobility data shows a larger drop followed by an increase in community mobility than was visible through our own data. 

One plausible explanation for this is the upwards sampling bias that occurs when using smartphone data \cite{BirenboimAmit2016MRit, ThimmTatjana2016Ppaf}. Our data set captures anyone on the street, including individuals experiencing homelessness who are less likely to have smartphones. This population was on the streets throughout the entirety of the pandemic, so they were consistently captured by our data collection efforts. This consistent baseline pedestrian count could lead to a lesser response to vaccine rollout and winter weather in our own data in comparison with Google's. Additionally, there is a known income gap in both vaccination rates and smartphone ownership \cite{BarryVaughn2021PiCV,CellData}. This gap could drive the increase in the Google Mobility data during vaccine rollout.

\subsection*{Implications, Limitations, and Extensions}
Our results show that it is possible for researchers to collect and analyze longitudinal SVI data. The presented methods can be used to collect and process SVI data from 8 hours worth of video in a manner of days. This time will only further decrease with faster data processing infrastructure and methods. These methods will allow novel longitudinal SVI data to be collected for research in a variety of application areas. 

The results of the case study also bear further discussion. We demonstrated expected relationships between seasonal effects like day of week and weather on pedestrian traffic. Additionally, we showed that pedestrian traffic is inversely proportional to income, a known result during the COVID-19 pandemic \cite{weill_social_2020, elarde_change_2021}. Our results also showed that more white areas had higher on average pedestrian counts. This could be due to known trends, such as areas with larger non-white populations being more likely to stay home in response to government restrictions \cite{singh_impacts_2021}, or just due to local trends, as racial mobility trends tend to vary between cities \cite{chang_mobility_2021}. These results validate our method with respect to established literature. 

One new finding from our case study is that while overall pedestrian counts did not respond to vaccine availability, the subset of pedestrians who were in larger groups (4+ people in an image) did. Likely, the reason we did not see a response to the vaccine in the aggregate data is because our data only captures people who are outdoors. There is  data that shows that outdoor pedestrian activity varied across cities, frequently increasing at recreation locations like trails, during the early days of the pandemic \cite{kraus_provisional_2021, doubleday_how_2021}. Given these increases at some locations, a return to 'normal' pedestrian traffic may not mean an increase, but rather a change in traffic patterns. Our data captures this by showing showing that there was a significant increase in larger groups of people after the vaccine became available. This implies that people were more willing to be near each other outdoors after they had been vaccinated. 

While the data pipeline presented here does represent a method for generating a novel data product, there are implementation challenges worth further discussion. For data collection, in addition to the time required to drive the route limiting the places of interest the route could reach, there were also many tradeoffs that had to be made when designing the route itself \cite{errett2023street}. Despite having our survey route carefully designed to assess a representative sample of the Seattle population, some bias in route design is unavoidable. Since the route design included data from the American Community Survey aggregated at the census tract level, there is an implicit assumption of spatial homogeneity of the population within each census tract. Such bias is a manifestation of the well-known modifiable areal unit problem \cite{fotheringham1991modifiable}. Since the majority of the route was primarily based on locations of interest throughout the city, this concern is somewhat mitigated. 

In terms of processing, the pre-trained model we used required a substantial amount of high-performance computing time, and at times the data product generated was so large as to be unwieldy. Given the challenge our data set represents, using a model designed to be generalizeable is necessary to attain good detection results. As many state-of-the-art models perform substantially worse out of sample, we had to be careful to choose a model that was designed to perform well in this situation, at the cost of slower computing times \cite{hasan2021pedestron}. Another unforeseen challenge was regular updates to the video camera's software to process and segment the video data into images. Consistent image formatting was vital for the data processing pipeline to function, so regular quality checks are necessary to make sure the images are processed properly.

The data product created, pedestrian detections, has some limitations as well. First, our method only captures pedestrians who are outdoors and near enough to the street to be captured via camera. This means that our data set does not include people who are indoors at these locations of interest, or who are too far from the street to be seen by camera. While the changes over time in pedestrian traffic we observed are still meaningful, it is important to recognize they don't capture everything. Similarly, our data cannot be interpreted as the actual number of pedestrians on the street. There is overlap in the image data, even when subset at 4 meter intervals and cropped during orthorectification. The orthorectified images only represent about 25\% of the originals. However, this natural cropping is not enough to avoid the image overlap and further cropping would risk information loss. Pedestrians that appear in the foreground of one image may end up in the background of another. There are also several known instances of cyclists keeping relative pace with the street-view vehicle for several blocks, resulting in numerous detections. These issues are easy to circumvent in analysis by comparing the relative number of detections, although at the cost of interpretability.

Even with the above limitations, the data pipeline presented in this paper can be directly applied or adapted to be used in a number of contexts. Potential applications of longitudinal SVI data in assessing the built environment \cite{smith_google_2021}, broad urban research \cite{cinnamon_panoramic_2021, biljecki_street_2021, li_street_2022}, and health research \cite{rzotkiewicz_systematic_2018} have been well-documented, as the temporal instability of existing SVI data is discussed as a limitation in all of these fields. Beyond this, it is possible to estimate population demographics \cite{GebruTimnit2017Udla}, and other neighborhood-level statistics \cite{suel_multimodal_2021,gullon_measuring_2023 } using SVI data. As our ability to quickly and accurately parse scenes using computer vision improves \cite{dong2022}, potential application areas will only increase in number. 

Another field where longitudinal SVI data could contribute a lot is disaster research. There is a substantial body of research dedicated to empirical methods for modeling various aspects of disaster recovery \cite{MartellMatthew2021RoEQ}. Our methods could be applied in this field to quantify recovery using pedestrian detections as a metric for community mobility, or another metric assessing the built environment as appropriate. Similar work has been done using repeat photography after Hurricane Katrina \cite{burton_evaluating_2011} but our methods represent a substantial increase in generated data, allowing for a wider range of analyses. Spatial video data collection for disaster reconnaissance has also been done \cite{curtis_spatial_2012}, but involves manual assessment of the captured video. Our methods demonstrate that a fully-automated approach is possible, which would allow for more frequent data collection at a lower cost.

\section*{Conclusion}

Regression analysis based on longitudinal SVI data showed that pedestrian traffic patterns changed in response to the availability of the COVID-19 vaccine. Our results demonstrate the feasibility and value in collecting SVI data as a part of a longitudinal study. Longitudinal SVI data is capable of providing valuable insights in a variety of fields of study.

\section*{Supporting information}


\paragraph*{S1 Table.}
\label{S1_Table}
{\bf OLS Regression Results for Detections per Image for the detections subset sharing an image with at least one other.} 

\paragraph*{S2 Table.}
\label{S2_Table}
{\bf OLS Regression Results for Detections per Image for the detections subset sharing an image with at least two others.} 

\paragraph*{S3 Table.}
\label{S3_Table}
{\bf OLS Regression Results for Detections per Image for the detections subset sharing an image with at least three others.} 

\paragraph*{S4 Dataset.}
\label{S4_Dataset}
{\bf Full dataset used for obtaining regression results presented in this paper.} 

\section*{Acknowledgments}
The authors gratefully acknowledge DesignSafe and the Texas Advanced Computing Center (TACC) at The University of Texas at Austin for providing the cyberinfrastructure that enabled the research results reported within this paper.

\nolinenumbers

%
%
%

\bibliography{main}

\end{document}